# Learning-Based Pressure Predictive Control of a Vertebraic Soft Robotic Tail

Wenjian Yang†, Nan Huang†, Yukang Nie, Fang Chen, Wanchao Chi*,
Jian S Dai, Fellow, IEEE, Sicong Liu*, Member, IEEE

***Abstract*—Soft robots have attracted much attention for their safe human-robot interaction and flexibility, but the typical continuum structure and nonlinear material behavior make the kinematics modelling complex, especially in non-static motions. In this work, we proposed an LSTM-based pressure predictive control (PPC) for the motion control of a vertebraic soft robotic tail and the coordination with a quadruped robot. The PPC consists of an inverse kinematics (IK) model, a forward kinematics (FK) model and a pressure compensation (P-comp) model, and achieves non-static and quasi-static motion control of the tail. Compared with the IK-only model, the average RMSE of the PPC's simulation trajectories reduces by 69.8%, when executing target trajectories. In the coordinated motions of the soft tail quadruped, using a prediction data set to train the PPC enables next-moment action prediction and reduces computation time by 60.9%, which enhances the real-time response of the tail to match the quadruped torso's moving rate. The PPC provides a simple and effective method to model the soft tail for both non-static and quasi-static motion control, and grants the soft tail quadruped with the functionality of interacting with the environment.**

## I. INTRODUCTION

Vertebrate tails with body-coordinated motions enhance animals' adaptability in unstructured environments. When a monkey is climbing, the tail can serve as an extra hand to wrap around branches at high speed or grasp objects in addition to maintaining body balance [1]. Inspired by vertebrate anatomy, researchers have applied robotic tails for quadruped robots to improve the performance in high-speed motion [2], [3]. The researches on robotic tails are mainly focused on motor-driven designs with rigid structures or cable-driven continuum structures [4]-[9]. The degrees of freedom, velocity and amplitude of these robotic tails are constrained by the number and load capacity of the motors. The safety of human-robot interaction during the movement of the tail is a concern due to the rigid structure. The off-center weight can also affect the robot's center of mass, impacting the motion stability during high-speed movements and further complicating modeling and control.

Continuum soft robots driven by pneumatic actuators show great potential for serving as the soft robotic tail due to the lightweight, compliance for adapting unstructured environments, and inherent safety in human-robot interaction. A typical design of such a robot consists of a series of soft joints. Each soft joint consists of several soft actuators and has up to 6 degrees of freedom [10], [11]. Soft origami actuator (SOA) [12], [13] made with simi-rigid soft materials, has been proved to be efficient to increase the stiffness, enhance the load capacity and response speed of the soft robot, showing the potential to achieve non-static motions in soft robots.

**Fig. 1.** The concept of PPC and the soft tail quadruped. (a) The target trajectory is put into PPC to generate pressure sequences, then the tail executes the predicted pressure to output the trajectory. (b) The BVSR tail is integrated with the quadruped robot to achieve coordinated motions and the capability to interact with the environment.

Due to the compliance and complexity in the soft robotic structures, the explicit theoretical modelling on the kinematics of such soft robots is challenging. The Denavit-Hartenberg (D-H) method based on the constant curvature assumption to build the kinematics model [14], [15] and the absolute nodal coordinate formulation (ANCF) modeling approach [16] need to comprehensively consider elasticity, external forces and geometrical constraints. The review [17] critically analyzes the complexity of modeling using physical model-based methods and the great advantages of being applicability to all system of the modeling using data-driven methods.

Without explicit analytical models, learning-based models for motion control show promising results in the soft robotic field. By collecting extensive experimental data, machine learning is used to construct numerical models that describe the kinematic and dynamic characteristics of soft robots [18], [19]. The method of long-term time series prediction based on nonlinear autoregressive networks with exogenous input (NARX) has been used to build kinematic and dynamic models [20]. Recurrent neural networks (RNNs) modeling methods attract increasing research interests for the control of soft robots [21], [22], [23]. Among the RNN methods, the frameworks for training the kinematic model of soft joints with bellow actuators based on Long Short-Term Memory (LSTM) [24] and Gated Recurrent Unit (GRU) [25] deep neural networks were proposed. They collected data from pressure sensors and a motion capture system to establish the mapping between pressures and positions. These works validate the feasibility of the learning-based method in quasi-static motions and control of the soft robots, and the LSTM network is proven to be simple and effective. The pressure sequence output by the inverse kinematics network can be directly utilized for the motion control of soft robots. However, challenges remain in the learning-based motion control of soft robots with non-static motions or the coordinated motions integrating with mobile robots.

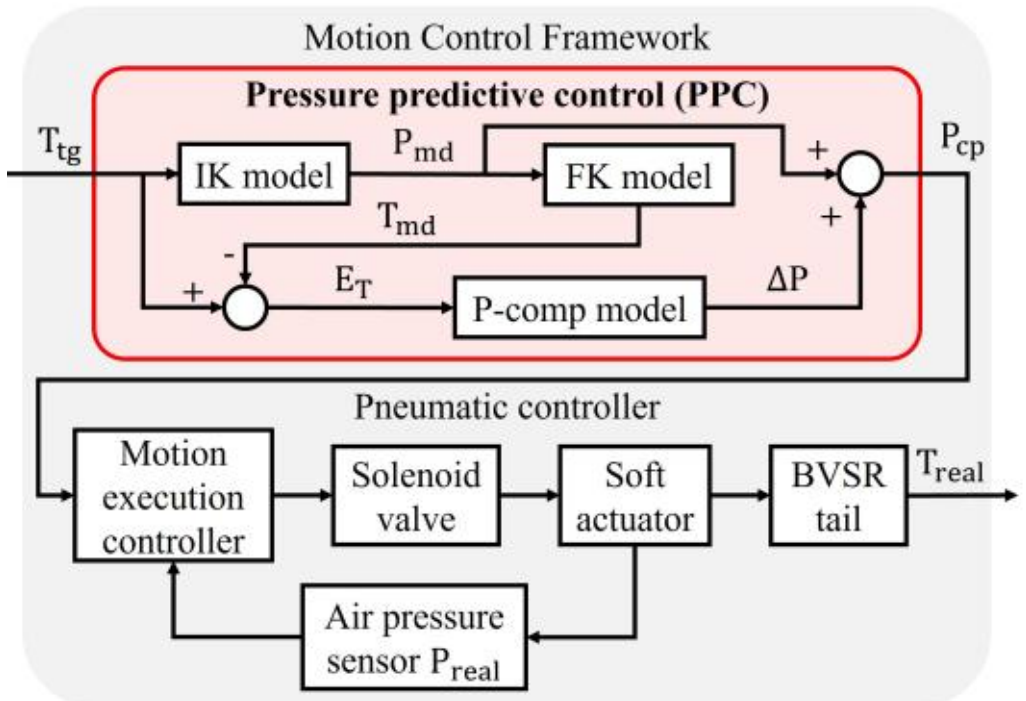

**Fig. 2**. Diagram of motion control framework of the BVSR tail. The proposed PPC includes IK, FK and P-comp models, the output pressure sequence can be directly executed by the pneumatic controller to drive the BVSR tail to output desired trajectory.

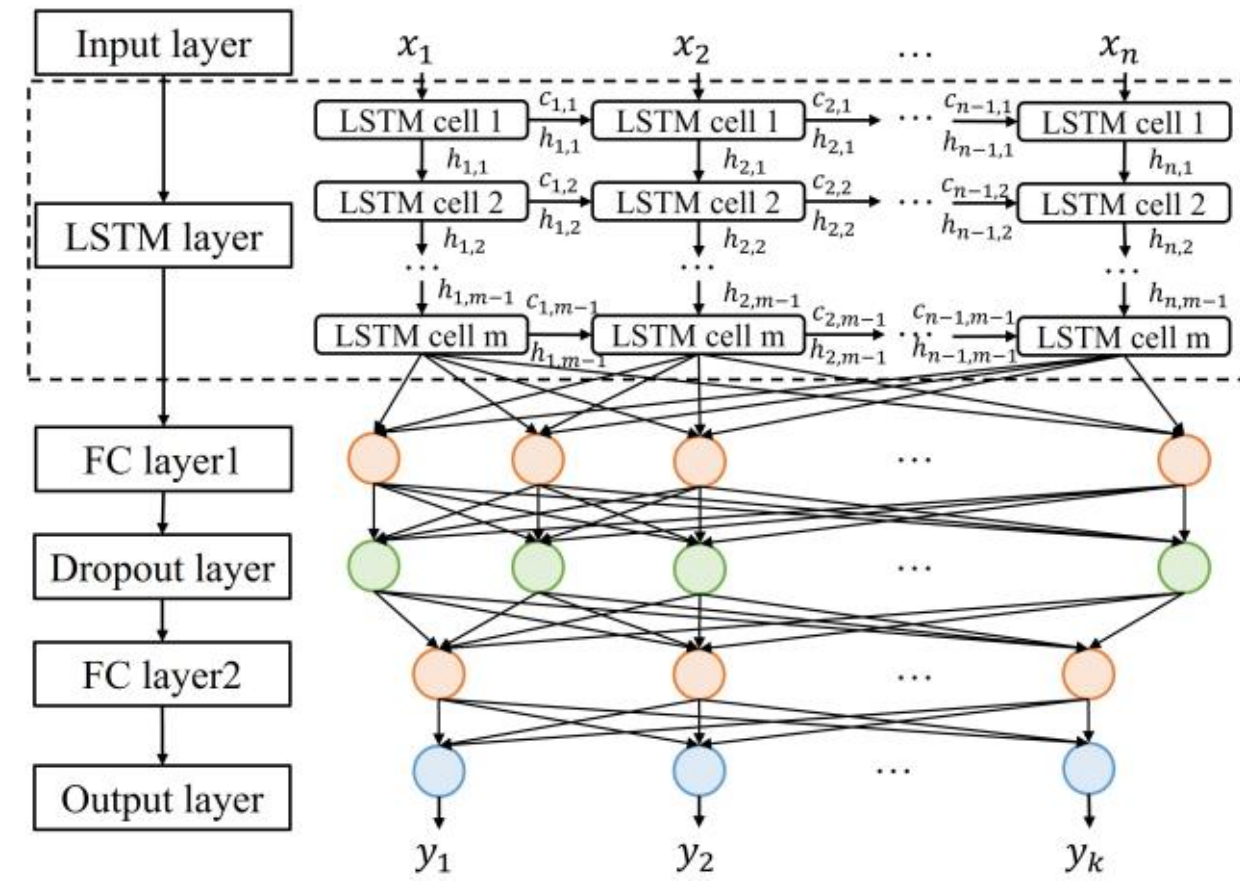

**Fig. 3.** Diagram of the LSTM network, including Input, LSTM, FC 1, Dropout, FC 2 and Output layer.

Although there have been numerous studies on mobile robots equipped with robotic arm, most of these studies have focused on rigid-body robots [4]-[9]. Research on the integration of soft tail with mobile platform remain scarce. The recent work [26] is among the few that integrate soft tails with mobile platforms, focusing on the tail's grasping and dynamic mobility enhancement to the quadruped. Among the exist studies, investigations on the motion accuracy of soft tail are lacking especially when performing non-static motions. Due to the inherent compliance of soft robots, the motion control often generates quasi-static trajectories which limits the applications. This work introduces an approach to address the challenge of the motion control of the soft robot with non-static motions and coordinated motions integrating with mobile robots.

In this paper, we propose a pressure predictive control (PPC) method based on LSTM network, aiming to alleviate the complexity in the kinematic modeling of soft robots consisting of continuum structures and nonlinear materials. The predicted air pressure sequence is directly applied to the motion of the biomimetic vertebraic soft robotic (BVSR) tail to output non-static and quasi-static movements as shown in Fig. 1(a). The tail is integrated on a quadruped robot and established coordinated motions as shown in Fig. 1(b). To reduce the delay of the coordinated motion and improve the real-time performance, the PPC is optimized, and hence the computation cost of the actuation system is reduced. The contributions of this work can be summarized as follows:

- An LSTM-based pressure predictive control (PPC) is proposed for motion control of both non-static (60 ~ 85 mm/s) and quasi-static (12 ~ 17 mm/s) motions of the BVSR tail. The PPC consists of an inverse kinematics (IK) network, a forward kinematics (FK) network, and a pressure compensation (P-comp) network, which are of the same LSTM architecture.
- The PPC is validated on the tail and compared with the IK-only model, which generates pressures, with averaged reduction of RMSE in trajectories by 69.8% (69.4% in Y, 69.8% in Z) in simulations, and by 18.8% (12% in Y, 24% in Z) in experiments, respectively.
- The PPC enables the tail and the quadruped with coordinated motions, and grants the soft tail quadruped with functionality for real-world application. Optimized by the IK-prediction (IKP) model, the PPC achieves the prediction of the tail's next action, reduces the computation cost and improves the real-time performance in coordinated motions.

This article is organized as follows: Section II presents the motion control framework with PPC. Section III presents the soft tail system and PPC implement. Validations are presented in Section IV, the implementation on the quadruped robot is included in Section V. The contributions, comparison and limitations are discussed in Section VI. Section VII concludes the paper.

## II. Motion Control Framework With PPC

### *A. LSTM-Based Pressure Predictive Control*

The pressure predictive control (PPC) consists of an inverse kinematics (IK) model, a forward kinematics (FK) model, and a pressure compensation (P-comp) model as shown in the motion control framework in Fig. 2. Due to the inherent hysteresis of pneumatic soft robots, there is a time disparity between the current air pressure and the current trajectory when performing non-static motions. This difference can be eliminated by using the P-comp network.

The input $T_i$ and output $P_i$ of PPC are illustrated below:

$$P_i = f(T_i, t), \tag{1}$$

where $P_i$ is the 10-channels air pressure sequence, the trajectory $T_i$ of tail's end-point Q is decomposed into the position coordinates $Y_i$ and $Z_i$ of the YOZ plane. The length of the sequence is denoted by $t$.

As shown in Fig. 2, the target trajectory $T_{tg}$ is processed by the inverse kinematics (IK) model to calculate the predicted input pressure values $P_{md}$. These pressure values are then fed into the forward kinematics (FK) model to obtain the predicted motion trajectory $T_{md}$. There is usually a discrepancy error $E_T$ between the predicted and target trajectory, which is used as inputs for the pressure compensation (P-comp) model. P-comp computes the necessary pressure correction values $\Delta P$ for the actuators to minimize the trajectory error. Finally, the corrected pressure values $P_{cp}$ are output to the pneumatic controller to drive the soft tail.

Fig. 3 illustrates the architecture of the LSTM network for IK, FK and P-comp, in which x represents the input sequence (e.g., continuous pressure or trajectory commands), and y denotes the final output sequence (e.g., predicted trajectories or pressure values). Within the LSTM unit, h is the hidden state responsible for transmitting short-term information, while c is the cell state that serves as a long-term memory carrier for retaining historically critical information. The network begins with a sequence input layer, followed by an LSTM layer which included sigmoid and tanh function as introduced in [27], followed by a fully connected layer (FC1) and a dropout layer with a dropout probability of 0.5. Sequentially, there is another fully connected layer (FC2) with a size equal to the output sequence dimension, followed by a regression output layer.

The number of hidden units in the LSTM layer, the size of the fully connected layer FC1, and the initial learning rate of the network are three critical parameters that determine the effectiveness of network training. Using the MATLAB Deep Learning Toolbox and Bayesian optimization, we identified an optimal set of network parameters, as shown in Table I. The computation is performed on an AMD Ryzen 7 4700G with Radeon Graphics 3.60 GHz and 32GB RAM.

TABLE I
PARAMETERS OF THE LSTM NETWORK

| Network | Hidden Units | Fully Connected Layer (FC1) | Initial Learning Rate |
|---|---|---|---|
| FK | 164 | 70 | 0.015 |
| IK | 160 | 196 | 0.015 |
| P-comp | 128 | 64 | 0.02 |

### B. Pneumatic Controller

The pneumatic controller executes the PPC generated pressure sequence $P_{cp}$. As shown in Fig. 2, the motion execution controller receives the desired pressure values for pneumatic channels and regulates the opening and closing of solenoid valves based on the current and desired actuator pressure values. When the current pressure exceeds the desired value, the negative pressure valve opens, and the positive pressure valve closes; conversely, when the current pressure is below the desired value, the negative pressure valve closes, and the positive pressure valve opens. Finally, the BVSR tail reaches the desired state and outputs the trajectory $T_{real}$ in real time.

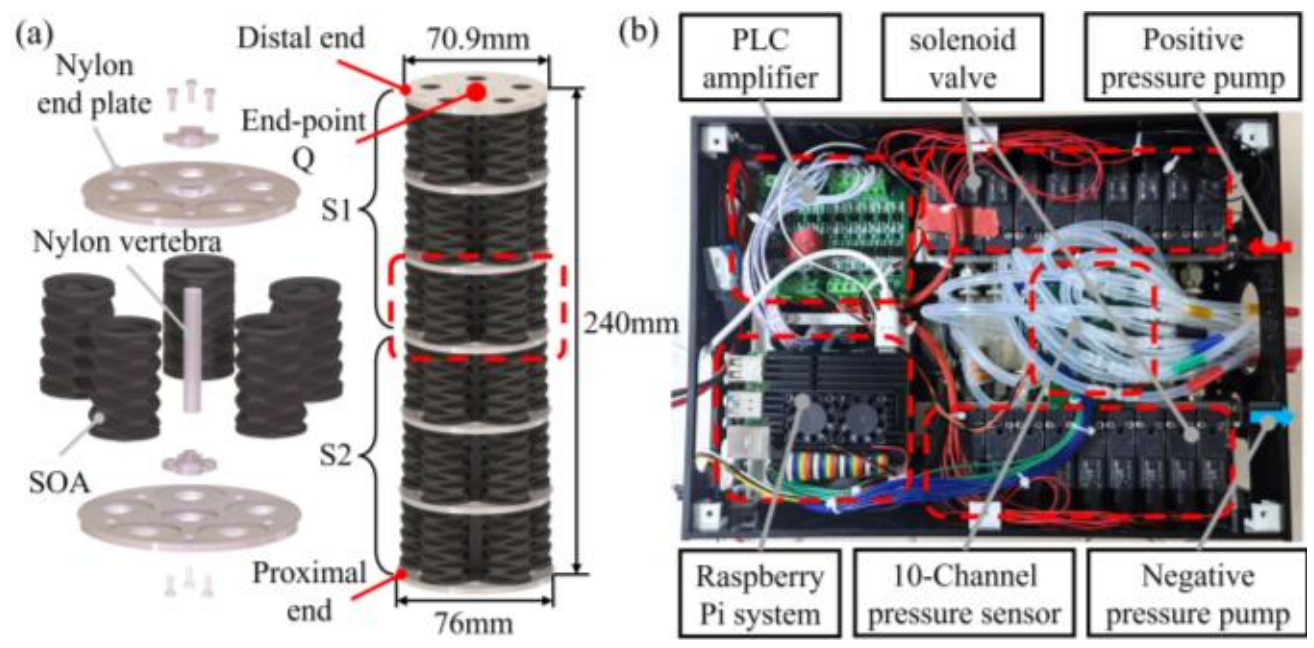

**Fig. 4.** The soft tail system. (a) The design of the BVSR tail. (b) The pneumatic control system.

## III. Soft Tail System and PPC Implement

### A. Design and Fabrication of the BVSR tail

Inspired by the tail of vertebrates, each joint of the BVSR tail adopts a biomimetic design by arranging the soft origami actuators (SOAs) as muscles around a Nylon elastic rod as the vertebra, achieving structure reinforcement and maintaining flexibility for non-static motions, as shown in Fig. 4(a). Each soft joint consists of 5 SOAs evenly arranged around the axis of the joint and glued both ends to the rigid end plates. The vertebra is fixed through the end plates at the axis. The origami actuators are made from polyurethane rubber through injection molding, while the rigid end plates are fabricated using PA12 Nylon via selective laser sintering (SLS) 3D printing. Given the higher Young's modulus of Nylon material compared to the polyurethane rubber, this design restricts the axial elongation of the soft joint while maintaining 2 degrees of freedom (2-DOF) bending, providing desired structural stiffness to enhance motion precision and stability. We then designed the BVSR tail by stacking 6 joints. Each set of three joints connected in series forms a segment (S1 or S2). In each segment, there are five independent pneumatic channels. The tail's length is 240 mm. The diameters of the proximal and distal ends are 76 mm and 70.9 mm, respectively. The weight is 222.4 g.

### B. Motion Controller and Sensing System

To drive the tail and obtain the pressure data, a pneumatic control system is designed as shown in Fig. 4(b). The BVSR tail is provided with pneumatic pressure by a positive pressure pump (Atlas Copco, 400 V, 4000 W, 200 L, 0.95 MPa) with a pressure regulating valve (SMC, 0.01-0.8 MPa), and a negative pressure pump (FUJIWARA, 220 V, 1500 W, 50 L). The positive and negative pressure pump is set to an absolute value of 160 kPa and 40 kPa, respectively, which are determined based on the operating range of the actuator (40 kPa ~ 210 kPa). The microcontroller is a Raspberry Pi (4B 4GB RAM) running the Linux system; it connects two external IO expansion modules (SZYTF PCF8575) by using Inter-Integrated Circuit (I2C) communication protocol. The control voltage is increased through a PLC amplifier (Macrowis, NPN) to drive 20-channel solenoid valves (OST Solenoid SY2/2N.C, 24 V, 2.5 W) for the pressurization and depressurization of 10 channels. The pressure value of each channel is measured by a 10-way pneumatic pressure sensor (Honeywell SSCDANN060PAAA5, 0 ~ 413.69 kPa), collected by a data acquisition module (ANFULAI AD7606) and sent to the Raspberry Pi by using Serial Peripheral Interface (SPI) communication protocol.

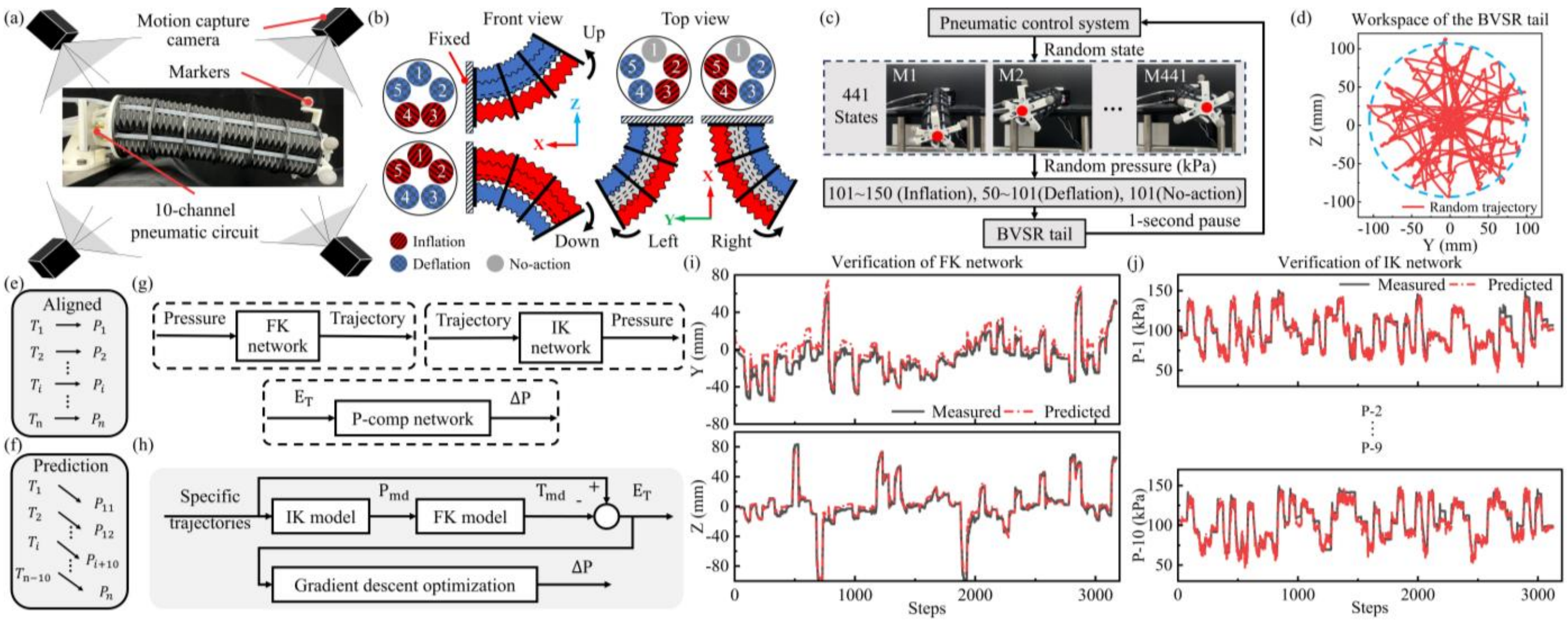

**Fig. 5.** Data acquisition and network training. (a) Experimental platform and motion capture setup. (b) Four basic states, bending around the Y-axis (up and down) and the Z-axis (left and right). (c) The BVSR tail randomly executes 441 states, with random pressure. (d) The workspace of the tail is obtained by the circumscribed circle of the random movement recordings. Formats of the data set are classified into (e) Aligned and (f) Prediction. (g) Training processes for FK, IK and P-comp network and (h) the process to obtain P-comp network training data. Verifications of (i) the FK network and (j) IK network.

As shown in Fig. 5(a), The tail is fixed horizontally, an array of motion-capture cameras identifies the five makers at the distal end of the tail, obtains the trajectory information, and sends the information to the host computer. The trajectory of the tail's end-point Q and the pressure data are used for network training.

### C. Data Acquisition

To collect data for network training, the soft robotic tail was driven to different positions under a series of different input pressure sets. The vertebral structure limits the axial movement of the tail. For a single segment, certain sets of input pressures cannot generate actual tail movement, such as inflating or deflating all actuators simultaneously. To enhance the usability of the collected data, this study designed four basic states as shown in Fig. 5(b), based on the segment structure and typical tail swinging states. The number of tail movement states S is determined by the following equation:

$$S = (S_{basic} * m + S_{initial})^n, \quad (2)$$

where $S_{basic} = 4$ represents the number of basic states, each state has $m = 5$ substates due to the rotational symmetry of the segment, $S_{initial} = 1$ denotes the initial state, and $n = 2$ indicates the number of segments. This configuration results in a total of 441 states for the tail, which are only used to determine whether each chamber is in inflation, deflation or no-action condition.

During data acquisition, as shown in Figure 5 (c), the tail randomly selects 441 states from S1 to S441 to define the condition of each chamber. Subsequently, the pressure for each chamber is randomly assigned within the following ranges: 101 kPa to 150 kPa for inflation condition, 50 kPa to 101 kPa for deflation, and 101 kPa for no-action. The tail cyclically executes this process in non-static motion, with each cycle interval of 1 s and repeated 60 times. The system completes the collection of a set of pressure and trajectory data at a frequency of 120 Hz, and 18 sets of data were recorded, yielding a total of 129,600 data pairs (60∗120∗18). To be consistent with the tail's actuation frequency and reduce data redundancy, the data were resampled at a frequency of 50 Hz with temporal synchronization between the pressure and trajectory. As a result, 48,000 pairs of data were collected, of which 80% and 20% of the data [19] were used as training set (15 sets) and test set (3 sets), respectively. Validation set is the same as the test set and is used to avoid overfitting in network training. Fig. 5(d) shows that a set of random trajectories in data acquisition process is fully covered by a blue workspace circle with the diameter of approximately 200 mm, which also proves the randomness of the data. The trajectory data $T_i$ ($i$ represent the time instant) of the end-point Q and the simultaneous pressure data $P_i$ of the actuators were recorded and used as an aligned data set for network training, as shown in Fig. 5(e).

### D. Network Training

Three networks need to be trained to form the PPC. To train the FK network and the IK network as shown in Fig. 5(g), the FK network accepts pressure sequences as input and outputs trajectories, while the IK network accepts trajectory inputs and outputs pressure sequences. The trajectories (pressure) generated by the FK (IK) network and the real trajectories (pressure) are shown in Fig. 5(i) (Fig. 5(j)). The training and test RMSE of the FK and IK network is shown in Table II. The result shows that the FK network provides effective prediction of the trajectory, with the average test RMSE (8.8 mm, 7.7 mm) in the form of (Y coordinates, Z coordinates). However, there is a deviation between the pressure predicted by the IK network and the measured pressure, with the average test RMSE 7.5 kPa. To improve the kinematic model's accuracy, it is necessary to use the P-comp network to compensate the output pressure. Thus, the FK and IK model for constituting the PPC are obtained and used to generate the training data set of P-comp network.

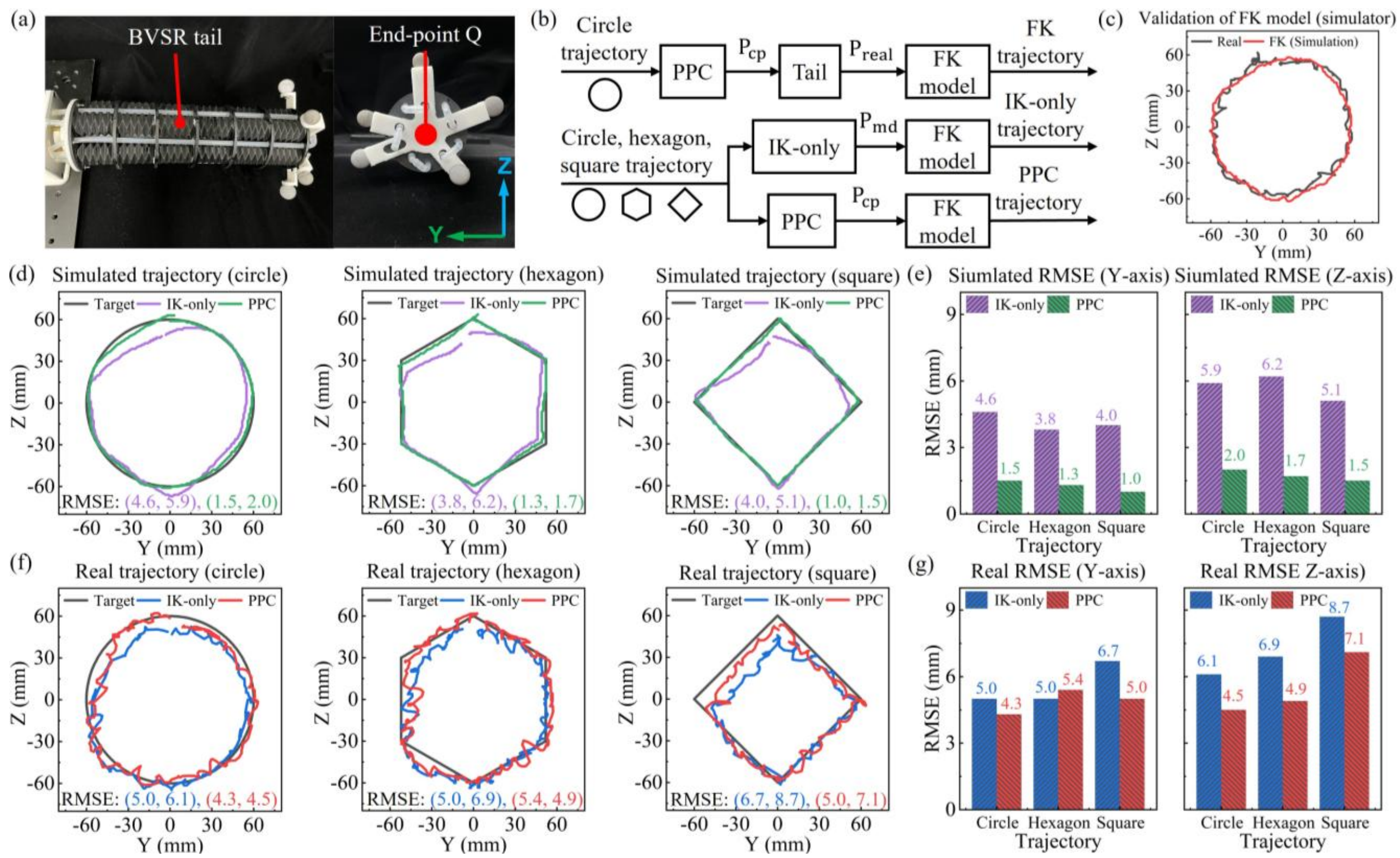

**Fig. 6.** Experimental verification of the non-static motions of the BVSR tail. (a) Experimental setup. (b) Processes for generating three different trajectories for comparison. (c) Validation of the FK model as the simulator for generating trajectories. (d) Comparisons of the PPC and IK-only simulated trajectories with the target trajectories and (e) their RMSE in Y and Z directions. (f) Comparisons of the PPC and IK-only experimental trajectories with the targets (g) their RMSE in Y and Z directions.

TABLE II
RMSE OF THE FK AND IK NETWORK

| Network | | Training | Test1 | Test2 | Test3 |
|---|---|---|---|---|---|
| FK | Y (mm) | 5.2 | 9.6 | 7.6 | 9.2 |
| | Z (mm) | 5.2 | 9.8 | 4.5 | 8.8 |
| IK | P (kPa) | 5.5 | 7.0 | 8.0 | 7.5 |

As shown in the Fig. 5(h), for the training of the P-comp network, the specific trajectories (circle, hexagon, square, etc. in total of 6,926 sequences trajectory data) were respectively put into the IK model to output $P_{md}$ and then put into the FK model to output $T_{md}$. The three specific trajectories were subtracted from the corresponding $T_{md}$ to obtain the error $E_T$. Since $\Delta P$ cannot be experimentally measured during network training, it is obtained through a gradient descent optimization method. This involves defining an appropriate loss function to measure the trajectory error, such as Mean Squared Error (MSE). $\Delta P$ is initiated for each sample as a parameter requiring gradient computation, and forward propagation is used to compute the current predicted trajectory and its error. Then, the gradient is calculated using backpropagation, which updates $\Delta P$ using an optimizer such as the Adam algorithm. This process is iterated until the loss function converges or a preset number of iterations is reached. By applying this method to multiple target trajectories, sets of pressure correction values $\Delta P$ are obtained, which are used together with $E_T$ to train the P-comp network to obtain the P-comp model. Specifically, the optimal pressure compensation sequence $\Delta P$ for a given target trajectory $T_{target}$ is obtained by solving the following optimization problem:

$$\min_{\Delta P} L(\Delta P) = \frac{1}{N}\sum_{i=1}^{N} \left|\left|FK\left(IK\left(T_{target},i\right) + \Delta Pi\right) - T_{target},i\right|\right|^2$$

$$s.t.\begin{cases}\Delta Pi \leq -50 \\ \Delta Pi \leq 50\end{cases}$$

where $\Delta Pi$ is the pressure compensation vector for the 10 pneumatic channels at the i-th timestep; $T_{target},i$ is the target coordinate at the i-th timestep; $IK(\cdot)$ and $FK(\cdot)$ are the pre-trained IK and FK models, respectively; N is the length of the trajectory sequence. The iteration terminates when the loss *L(ΔP)* falls below a threshold of $10^{-4}$ or after 100 epoch. Since $\Delta P$ and $E_T$ cannot be measured by experimental method, the verification of the P-comp network is included in the verification of the PPC in the next section. Finally, the IK model, the FK model and the P-comp model formed the PPC. By running the network training module only once, the models within PPC can be obtained.

# IV. Validation of the Motion Control Method

Here, we compared the trajectories of the PPC and non-PPC (IK-only) in simulation and experimental environments, respectively. The FK model is used as a simulator to generate simulated trajectories, and the experimental trajectories are obtained by the motion capture system.

## *A. Validation of Forward Kinematics Model*

In the experiment, we fixed the proximal end of the BVSR tail and left the distal end hanging like a cantilevered structure, as shown in Fig. 6(a). The pneumatic control system executes the pressure sequences at frequence 50 Hz. The real-time pressure $P_{real}$ and trajectory $T_{real}$ were recorded by pneumatic sensors and motion capture system, respectively.

To verify the obtained FK model in non-static motions prediction, we used a circular target trajectory with a radius of 60 mm and obtained a set of the Y-axis and Z-axis coordinate values corresponding to the points on the trajectory through the interpolation method. As shown in Fig. 6(b), pressure sequences $P_{cp}$ is calculated by the PPC from the target trajectory and sent to the tail. The tail executed in three repetitions with non-static motions at the end-point speed 75 mm/s. The real pressure $P_{real}$ are put into the FK model to generate FK trajectory, which is compared with the real trajectory as shown in Fig. 6(c). The FK trajectory shows a high degree of coincidence with the real trajectory and the average RMSE of the three experiments is (5.2 mm, 6.2 mm), which proves that the FK model can be used as a simulator to generate trajectories corresponding to the given pressure sequences.

## *B. Validation of PPC*

To verify the improvement of control accuracy by the PPC, it is compared with the IK-only. There are three typical shapes (circle, hexagon and square) as the target trajectories, which are completed by the tail with non-static motions (60 ~ 85 mm/s). The radius of the circle and the circumscribed circle of the hexagon and square are 60 mm. As shown in Fig. 6(b), the target trajectories are sent to the PPC and IK-only to derive the pressure sequences $P_{cp}$ and $P_{md}$, respectively. Then, these two pressure sequences are put into the simulator (FK model) to generate simulated trajectories, as shown in Fig. 6(d). The RMSE of the simulated trajectory and the target trajectory is shown in Fig. 6(e) in the form of Y-axis and Z-axis.

The results show that the PPC trajectories are closer to the target than the IK-only trajectories in three cases. The RMSE of the three PPC trajectories are much lower than the IK-only trajectories. The average RMSE of the trajectories is reduced by 69.8% (69.4% in the Y-axis, 69.8% in the Z-axis coordinates) in simulation. Thus, by calculations, the PPC, i.e. the pressure predictive control with pressure compensation, is verified to be more accurate than the IK-only.

To verify the performance of the PPC in experiments, the IK-only generated $P_{md}$ and the PPC generated $P_{cp}$ pressure sequences of the three trajectories are sent to the pneumatic control system to execute on the BVSR tail. The real trajectories of the end-point are shown in Fig. 6(f), which show similar shapes as the targets as expected. As shown in Fig. 6(g), the RMSE between the real and the target trajectories of the PPC and IK-only are calculated and listed in Table III. The results show that the maximum RMSE of the real trajectory of PPC is 7.1 mm in both axes, which is 3.6% of the workspace diameter (200 mm) and is generally lower than the RMSE of IK-only. The average RMSE of trajectories is reduced by 18.8% (12% in Y, 24% in Z), which confirms that PPC trajectories are closer to the targets. Frequent switching of the solenoid valves causes intermittent impacts of the airflows in the pneumatic channels of the tail, and due to the inherent hysteresis of the soft material, the experimental trajectories display the jitter phenomenon. Due to the jitter, the RMSE of the real trajectories are higher than that of the simulated trajectories. Despite of the jitter, the PPC improves the accuracy of the experimental trajectories as the simulated.

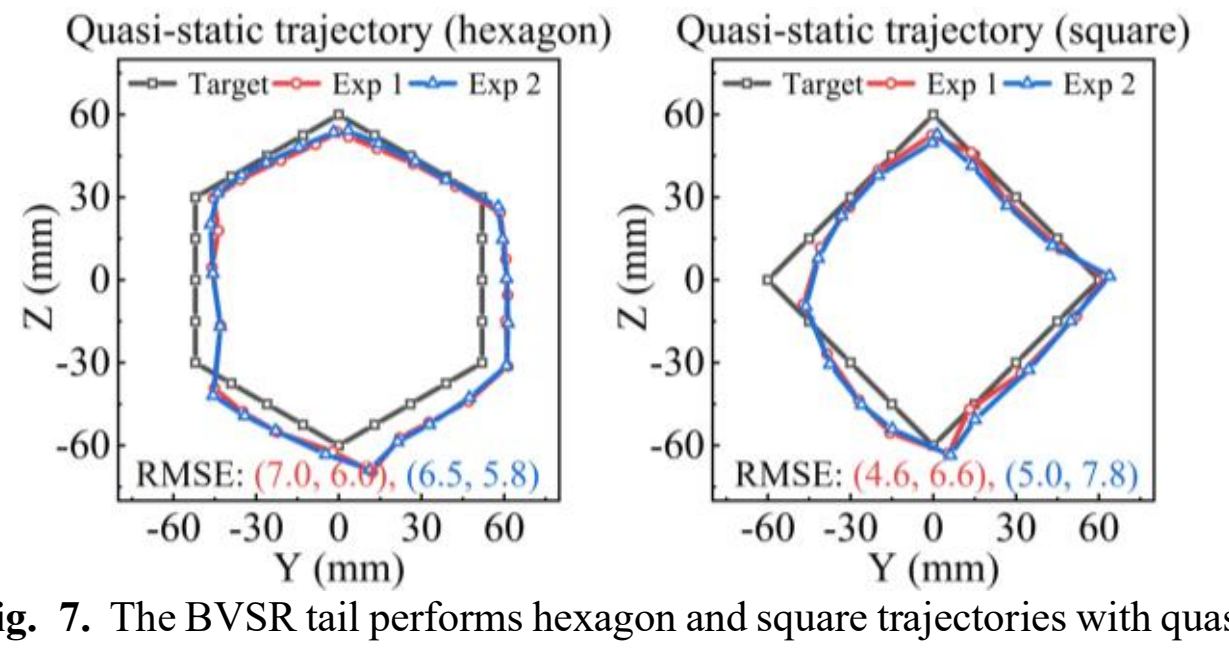

**Fig. 7.** The BVSR tail performs hexagon and square trajectories with quasi-static motions.

TABLE III
RMSE on the Y and Z-axis of PPC and IK-only

| Trajectory | Simulation (Y, Z) (mm) | | Experiment (Y, Z) (mm) | |
|---|---|---|---|---|
| | IK-only | PPC | IK-only | PPC |
| circle | (4.6, 5.9) | (1.5, 2.0) | (5.0, 6.1) | (4.3, 4.5) |
| hexagon | (3.8, 6.2) | (1.3, 1.7) | (5.0, 6.9) | (5.4, 4.9) |
| square | (4.0. 5.1) | (1.0, 1.5) | (6.7, 8.7) | (5.0, 7.1) |

## *C. Quasi-static Motion with PPC*

Apart from the non-static performance, the PPC is able to generate pressure sequences of quasi-static motions for the tail to perform, which is essential for human-robot interactions in the potential applications. In the quasi-static motions' cases, the BVSR tail completes the experimental trajectories at a lower speed range of 12 mm/s to 17 mm/s. The PPC produces the pressure sequences according to the trajectories, and the BVSR tail executes hexagon and square trajectories with quasi-static motions. Three points are interpolated on each edge of the polygons. During the movement, the tail stayed at each point for 1 second with the real coordinates recorded. We repeat the experiment twice in each trajectory. As shown in Fig. 7, the experimental recordings show similar polygonal shapes as the target trajectory with low RMSE. To compared with the quasi-static method in [24], the RMSE in the Y and Z are superimposed, thus the average RMSE of the hexagon and the square are 12.65 mm and 12 mm, respectively, while the RMSE for the square trajectories in [24] are 11.53 mm and 14.05 mm with 7 mm/s movements. Although the PPC was trained using non-static motions data, the similar RMSE verify the PPC can generate pressures for quasi-static motions, validating the adaptivity of PPC.

### *D. Validation of Network*

To systematically evaluate the performance differences between LSTM and GRU in modeling the motion of soft robots, this study trained and validated both architectures on identical tasks and datasets. Specifically, based on the IK-only framework, we employed LSTM and GRU networks for training respectively, and used the established FK to simulate trajectory predictions based on their outputs. As illustrated in Fig. 8, across multiple trajectory-tracking tasks, the IK-only model based on LSTM exhibited significantly lower trajectory error along the Y-axis compared to the GRU version, reflected in a reduce RMSE. This result validates the advantage of LSTM in capturing the motion characteristics of soft robots and provides an experimental basis for the selection of the training algorithm used in the PPC framework within this study.

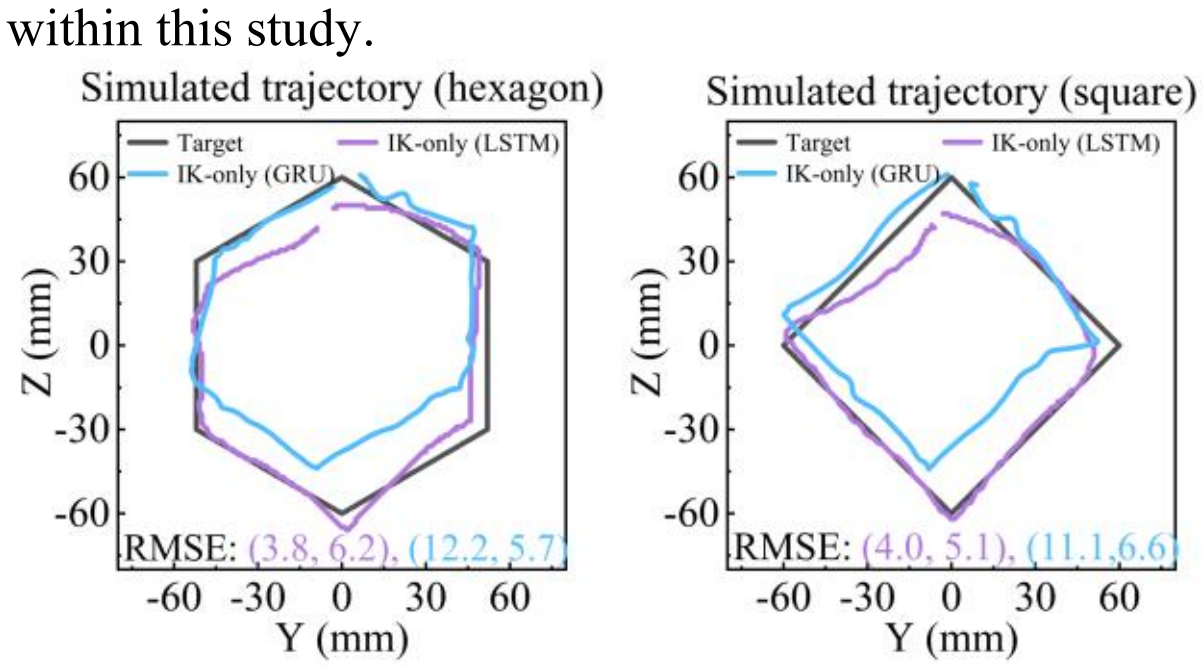

**Fig. 8.** IK-only performs hexagon and square trajectories based LSTM and GRU

### *E. Hysteresis Reduction*

During non-static motions, the inherent hysteresis characteristics of soft robots lead to a lagging displacement response behind the pressure command. The introduction of the P-comp module is intended to compensate for this phenomenon. As shown in Fig. 9, to quantitatively validate the inhibitory effect of PPC on hysteresis, we designed reciprocating motion experiments in both the horizontal and vertical directions. The hardware setup for the experiment is shown in Fig. 9(a). An inertial measurement unit (IMU) receives the attitude data of the BVSR's distal end in real-time at a frequency of 100 Hz, and the displacement of the end point is calculated.

The experimental procedures are as follows: based on the IK model and the PPC framework, we generated pressure command sequences corresponding to the horizontal as shown in Fig. 9(b) (target displacement range: -35~20 mm) and vertical as shown in Fig. 9(c) (target displacement range: -25~30 mm) reciprocating motions, and input them into the soft tail system for execution; each experiment was repeated three times, and the actual pressure and displacement data were recorded synchronously. By averaging and fitting the data from the three cycles, we obtained their respective hysteresis loops, and used the hysteresis rate as a quantitative indicator to evaluate the hysteresis performance of the models. In the horizontal motion, the hysteresis rate of the IK-only model was 22.27%, while PPC reduced it to 18.97%; in the vertical motion, the hysteresis rate of IK-only was 24.89%, and PPC further reduced it to 18.71%. This result indicates that PPC achieved a significant reduction in the hysteresis rate, thereby verifying the effectiveness of the P-comp module in mitigating the system's hysteresis effects.

### *F. Control Strategies Comparison*

To systematically evaluate the performance of the PPC in the field of soft robot control, a comparison has been conducted with representative control methods reported in relevant studies over the past five years. These methods include both physics-based control strategies [16] and data-driven approaches utilizing other neural network architectures, such as the Nonlinear Auto Regressive Exogenous (NARX) model [20] and Gated Recurrent Units (GRU) [25]. A comprehensive analysis was carried out with a series of metrices, including number of segments of the soft mechanical structure, trajectory tracking accuracy (represented by both RMSE and normalized RMSE relative to the workspace size) and motion velocity of the soft mechanical structure. The comparison results are presented in Table IV. It can be observed that the proposed soft tail system has the most segments and channel inputs. The horizontal placement is subjected to significant influence of gravity, which generally poses challenges for motion control. Despite these characters, the PPC demonstrates notable advantages in both absolute and normalized RMSE, underscoring its trajectory tracking precision. Furthermore, the PPC maintains satisfactory control performance across a broad range of motion velocities—from quasi-static to non-static regimes—highlighting its excellent motion adaptability.

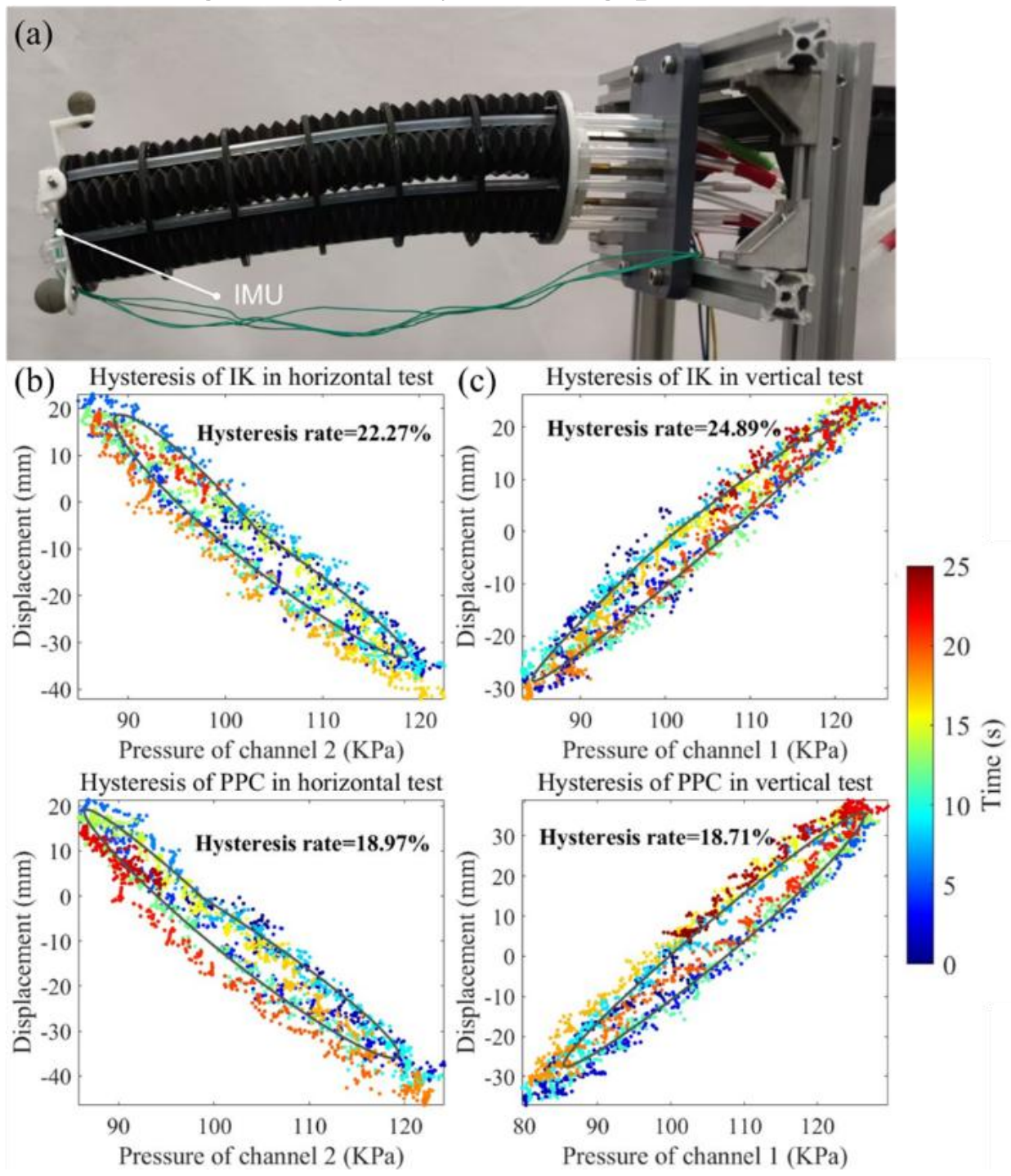

**Fig. 9.** Hysteresis comparisons between IK and PPC. (a) Horizontal test. (b) Vertical test.

## V. Motion of the Soft Tail Quadruped Robot

For the implementation on the mobile robotic platform, we adopt a quadruped robot to investigate coordinated motion between the tail and platform.

### A. Modelling of End-point Position

To test the coordinated behavior, the BVSR tail was installed on the rear end of the quadruped robot platform (Tencent Robotics X MAX6) which can be simplified as shown in Fig. 10(a). The quadruped robot can directly obtain the position $o_1$ of its center of mass (COM) and the Euler angle of its torso. The proximal end of the tail is fixed to the rear of the quadruped, so the coordinates of the tail's root $\boldsymbol{o_2}$ can be expressed as $\boldsymbol{o_2} = \boldsymbol{T_1}\boldsymbol{R_1}\boldsymbol{o_1}$ , where $\boldsymbol{T_1}$ is the translation matrix from COM to the BVSR tail's root. Rotation matrices $\boldsymbol{R_1} = \boldsymbol{R_x}\boldsymbol{R_y}\boldsymbol{R_z}$ can be used to represent the changes in rigid body angles, where $\boldsymbol{R_x}$, $\boldsymbol{R_y}$ and $\boldsymbol{R_z}$ are the rotation matrices around three axes, respectively. When the BVSR tail is inactivated, it can be regarded as a rigid cylinder to calculate the position of end-point Q, i.e. $\boldsymbol{q} = \boldsymbol{T_2}\boldsymbol{o_2}$, where $\boldsymbol{T_2}$ is the translation matrix from $\boldsymbol{o_2}$ to Q. In this way, position of Q can be obtained during the movements of the quadruped.

### B. Coordinated Motion Implementation

In the coordinated motions of the tail following the quadruped robot, the tail base angle changes with the pitch motions of the torso (-13° ~ 13°). It is necessary to study the influence on the output trajectories of the tail. The tail is mounted at three different base angles (-15°, 0°, 15°) to perform the same pressure sequence. As shown in the Fig. 10(b), the RMSE of the base angle at 15° and -15° is (3.9, 5.6) and (4.7, 5.9), respectively, which are very close to that of the horizontal state (0°) (3.9, 5.1). This shows the base angle does not affect the output trajectory obviously, which proves the training data collected in the horizontal state is applicable to the states with different torso pitch angles.

Moreover, the real-time performance of the tail should be considered to match the response rate of the quadruped motion. To obtained desired real-time performance from PPC, we adjusted the training data of the IK network by shifting the correspondence of the trajectory and pressure data, where the position $T_i$ is corresponding to the pressure at the $i + 10$ time instant $P_{i+10}$, as shown in Fig. 5(f). Using the new prediction data set, the IK-prediction (IKP) model is obtained and replaces the IK model in PPC, which realizes the prediction of the pressure at the next moment. The IKP network is verified using the test set, resulting in an RMSE of 10.3 kPa (in the pressure range of 50 ~ 150 kPa) that is slightly higher (2.8% relative increase) than the RMSE 7.5 kPa of the IK network. The small accuracy lose is an acceptable tradeoff for real-time performance enhancement.

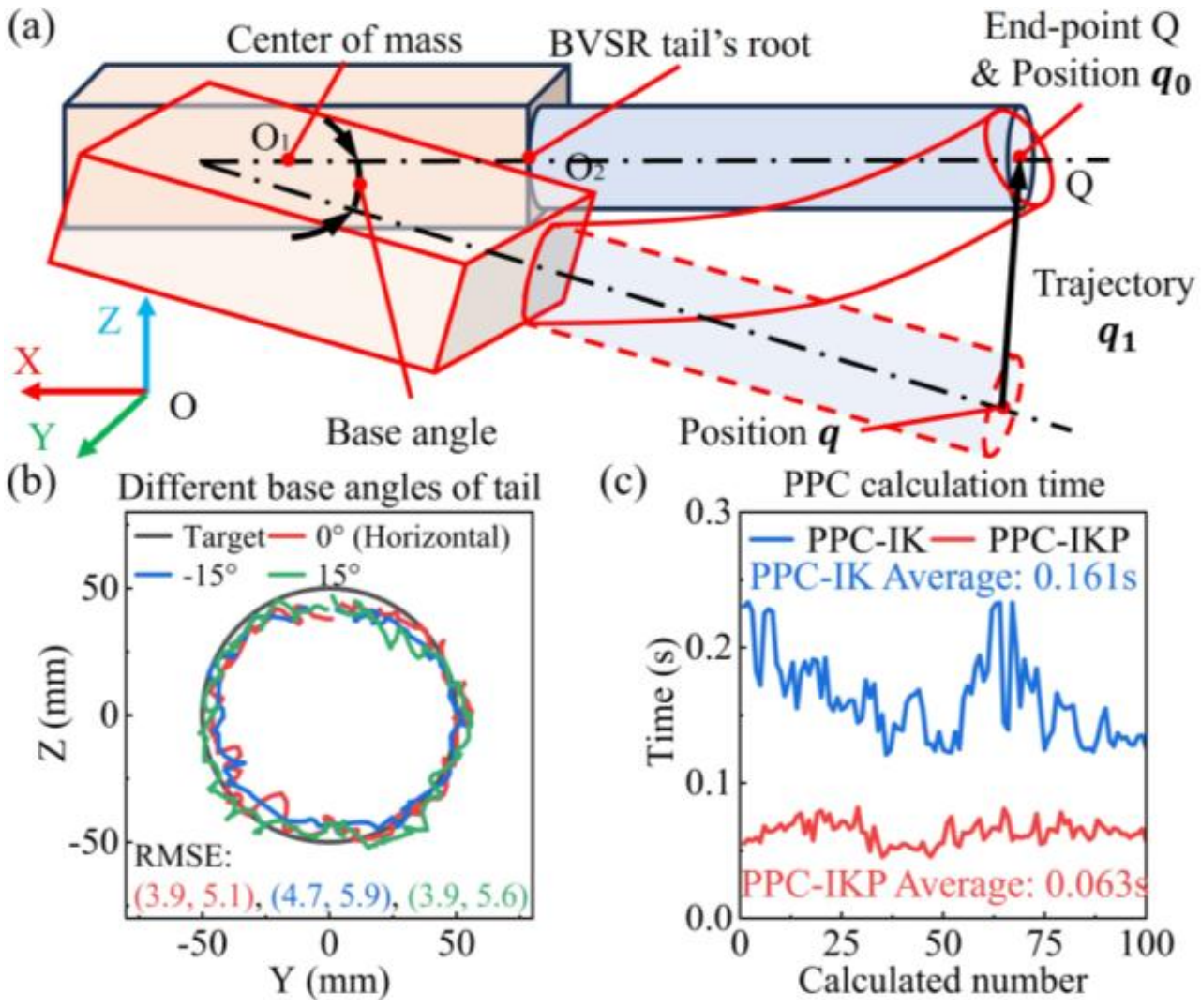

**Fig. 10.** Tail-robot coordinated motion implementation. (a) The orange block in the simplified model represents the quadruped torso, and the blue block represents the BVSR tail. (b) The tail performs the same pressure sequence at different base angles. The RMSE of each case corresponds to the curve with the same color. (c) Comparison of computation time between the PPC with IK model (PPC-IK) and IKP model (PPC-IKP).

The Raspberry Pi receives trajectory data at a frequency of 50 Hz from the quadruped, 100 sequences historical trajectory data ($T_{i-99} \sim T_i$) are put into the IKP model, but only the predicted pressure ($P_{i+1} \sim P_{i+10}$) are input to the FK model and P-comp model to complete the pressure compensation. This process reduces the average computation time from 0.161 s to 0.063 s as shown in Fig. 10(c), obtaining a 60.9% reduction. The real-time-performance-optimized PPC is hence used in the coordinated motion control of the soft tail quadruped.

### C. Tail End-point Control

Here, to validate the coordinated motion of the tail and quadruped, an end-point control experiment is setup where the tail aims to maintain the initial position of end-point Q while the robot moves its torso in a standing state, as shown in Fig. 10(a). When the quadruped robot is in the initial standing state, Q is at the initial position $\boldsymbol{q_0}$. Q will change as the quadruped robot twists its body in Yaw and Pitch directions, and by controlling the tail's movement accordingly, Q can be kept still at $\boldsymbol{q_0}$. Due to the movements of the torso, the change in the position of the tail should be compensated by the counter movement trajectory $\boldsymbol{q_1}$ of the tail, which is the trajectory the tail needs to move to keep the Q at $\boldsymbol{q_0}$, thus $\boldsymbol{q_1} = \boldsymbol{q_0} - \boldsymbol{q}$. The above process is calculated on the quadruped MAX6 (PICO-WHU4, 8th Gen Intel Core i7 Celeron Processor SoC).

The tail is installed at the rear of the quadruped robot and the pneumatic control box is on the back as shown in Fig. 11(a). First, we conduct experiments to investigate the moving range of Q, where the torse moves in Yaw direction with different speeds and the tail moves in response to keep Q at initial position. The results in Fig. 11(c) show that as the twisting speed increases, the Q's moving range became larger, i.e. the end-point Q become less stable. To investigate the capability of controlling the end point with Yaw and Pitch torso motions, the torso is programed to twist sinusoidally at speed of 2.6 °/s in the range of -13° to 13° in Yaw and Pitch directions for three times, while the tail moves in response to the torso as shown in Fig. 11(a) and (b). The real trajectories of Q are captured to compare with virtual trajectories $\boldsymbol{q}$ where the tail is inactive. Due to the inherent hysteresis of the soft tail and the computation time of the PPC, there are a small delay in the response movements, which makes the end-point lie in a small range of distance from $\boldsymbol{q_0}$ as indicated by the real trajectory curves in Fig. 11(d).

The recorded data are analyzed using RMSE, standard deviation (SD) and average tracking error on the Y-axis and Z-axis as listed in Table V. The ratio of the average RMSE over the virtual workspace is 9.3% in Y-axis (workspace, -101.0 ~ 107.0 mm) and 9.7% in Z-axis (workspace, -92.6 ~ 111 mm). We use SD to describe the degree of dispersion from the average distance, and to show the stability of Q. Compared to the size of the tail's distal end (70.9 mm), the average SD values are 26.3% (Y) and 22.6% (Z). The ratio of the RMSE and the SD values in the experiments indicates that the end-point position can be constantly maintained within a relatively small range. The average tracking error is -4.6 mm in the Y-axis, and -2.3 mm in the Z-axis, indicating the motion accuracy. In addition to the torso moving in two directions independently, the quadruped performs circular movements to further verify the ability to keep Q stable. As shown in Fig. 11(e), Q is kept in the area with maximum distances 50.9 mm in Y and 73.3 mm in Z from the target point, which are 23.3% and 28.8% over the motion ranges in Y (-117.6 ~ 100.7 mm) and Z (-142.5 ~ 111.8 mm), respectively. The experiments show that the tail achieves coordinated motions in response to the quadruped with desired real-time performance, under the control of PPC.

### *D. Bucket-picking Demonstration*

To show the functionality of the soft tail quadruped, an environment interaction demonstration is setup for the robot to pick up a bucket on the ground as shown in Fig. 11(f). The robot is firstly remotely controlled to approach the target location, and keeping the standing state. To reach for the bucket, the torso automatically pitches while the BVSR tail swings simultaneously following the torso motion, so that the reaching distance of tail's end-point Q is extended. Finally, Q reaches the pre-set target point, and the hook picks up the bucket. Fig. 11(g) shows the real trajectory of the Q generated by the combined motions of the tail's swinging and the torso's twisting, and shows the pitch angle of the torso. The torso twists -11.4° to facilitate the movement of Q to reach a total distance of 169.9 mm for successfully picking up the bucket. Unlike the previous experiment, here the coordination of the torso and tail extends the moving range of Q, which verifies the effectiveness of the PPC and demonstrates the functionality of the soft tail quadruped.

## VI. DISCUSSION

### *A. Interpretation of Key Contributions*

Experimental results demonstrate that the proposed neural network-based PPC framework delivers excellent control performance across different motion modes. During non-static motion (60–85 mm/s), it achieves a normalized RMSE ranging from 2.3% to 3.9%. Under quasi-static motion (12–17 mm/s), the RMSE is further reduced to between 2.15% and 3.5%. These results validate the high control precision of the PPC framework across varying speed ranges. Furthermore, in coordinated control experiments involving the soft tail and the quadruped robot, the PPC effectively mitigated disturbances caused by system dynamic coupling and maintained endpoint trajectory stability. This preliminarily confirms the framework's potential for enabling reliable interaction between the soft tail and the external environment when mounted on a dynamic base. Collectively, these experimental findings indicate that this study has not only made a breakthrough in improving the trajectory control accuracy of soft robots but has also achieved progress in effectively integrating soft robots with mobile platforms for practical application.

TABLE IV
THE COMPARISON BETWEEN THE PPC AND EXISTING CONTROL METHODS

| Studies (Year) | Basic algorithm | Number of segments | Placement state | RMSE range (mm) | Normalized RMSE range | Speed range(mm/s) |
|---|---|---|---|---|---|---|
| Huang et al. (2022) [16] | Physics based | 1(3 channels) | Vertical | 6.2~9.4 | 2.6%~3.9% | 10~40 |
| Thuruthel et al. (2018) [20] | NARX | 2(6 channels) | Vertical | 20.1~49.3 | 6.7~17% | - |
| Huang et al. (2024) [24] | LSTM | 1(3 channels) | Vertical | 10.78~14.15 | 3.6~4.39% | Quasi-static |
| Wang et al. (2023) [25] | LSTM | 1(3 channels) | Vertical | 1.08~1.12 | 6.35%~6.59% | Quasi-static (<13.3) |
| | GRU | | Vertical | 1.09~1.13 | 6.41%~6.65% | |
| **PPC (This work)** | **LSTM** | **2(10 channels)** | **Horizontal** | **4.3~7.1** | **2.15%~3.5%** | **Quasi-static (12~17)** |
| | | | | **4.6~7.8** | **2.3%~3.9%** | **Non-static (60~85)** |

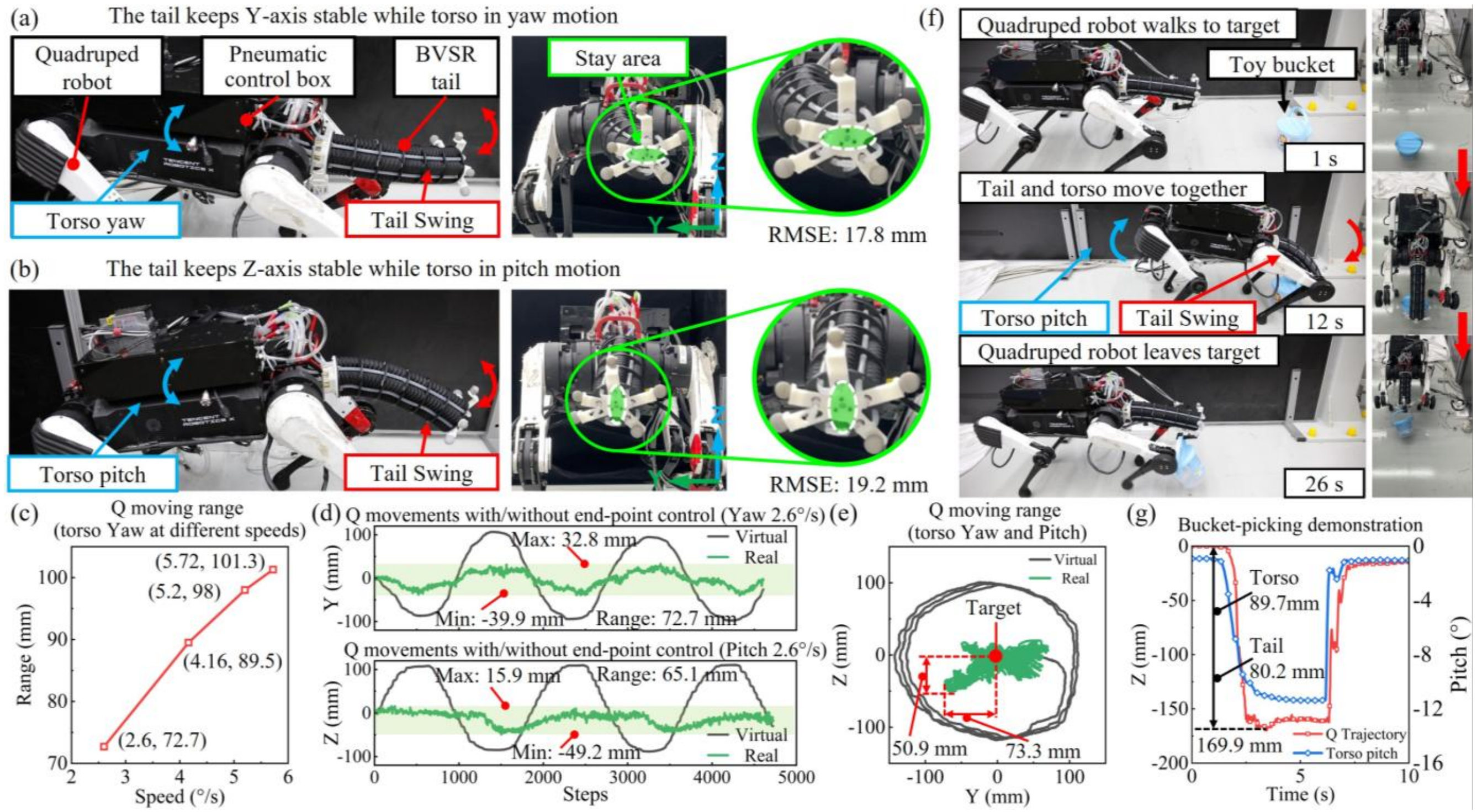

**Fig. 11.** Experiments on the soft tail quadruped coordinated motions. (a) The soft tail quadruped is equipped with control box and performs yaw and (b) pitch motion, with the tail swings in the Y and Z direction, respectively. (c) The moving ranges of Q when the torso twists at different speeds are listed as (speed, range). (d) The comparison between the virtual trajectory with the inactive tail and the real trajectory with the tail performing end-point control in the Y and the Z direction. The green area marks the displacement boundary of Q. (e) The torso moves in both yaw and pitch directions, Q is kept in the aera with maximum distances 73.3 mm in Y and 50.9 mm in Z from the target point. (f) By completing the bucket-picking task, the soft tail quadruped shows the preliminary capability to interact with the environment granted by the coordinated motion. (g) The Q's movement in Z and the torso pitch angle are recorded.

TABLE V
EVALUATION METRICS OF THE END-POINT CONTROL

| Direction | Y-axis (Yaw) (mm) | | | Z-axis (Pitch) (mm) | | |
|---|---|---|---|---|---|---|
| Experiment No. | 1 | 2 | 3 | 1 | 2 | 3 |
| RMSE | 17.8 | 19.4 | 20.7 | 19.3 | 20.5 | 19.2 |
| SD | 17.2 | 19.4 | 19.3 | 18.8 | 15.1 | 14.1 |
| Tracking error | -4.7 | -1.6 | -7.4 | -4.2 | -1.4 | -1.3 |

### *B. Comparison with the Soft Robotic Control Strategies*

As shown in Table IV, the proposed PPC framework in this study demonstrates promising performance even under high control complexity. Specifically, the soft-tail system controlled by PPC possesses the largest number of control channels (10 pneumatic inputs), and its horizontal configuration is subject to substantial gravitational influence, both of which increase the control difficulty. Nevertheless, compared to the physics-based control method [16] and the neural network-based control methods [20], [24], PPC achieves a lower RMSE in trajectory tracking; when compared to [25], PPC also exhibits higher accuracy control capability within the normalized RMSE range. Furthermore, the PPC framework shows a significant advantage in motion speed adaptability, maintaining control accuracy over a wide speed range, whereas existing methods are mostly confined to narrower speed intervals. In summary, the PPC method not only improves the trajectory control accuracy of the soft robot but also extends the applicability in non-static scenarios.

### *C. Limitation and Future Directions*

Although promising progress achieved in this study, several limitations warrant further exploration. Firstly, the proposed control algorithm has not yet fully accounted for the motion compensation of the soft robot under external loads or disturbances, and its performance evaluation is currently primarily confined to accuracy analysis under no-load and no-disturbance conditions. Considering the loads or disturbances conditions is set in the future stage to make PPC applicable to real-life tasks. Secondly, the coordinated motion achieved between the soft tail and the torso remains relatively rudimentary, showing a significant gap compared to the adaptive coordination capabilities exhibited by animals in nature. Thirdly, the motion speed of the tail was limited to 85 mm/s, due to the limitation arose from pneumatic hardware. At speeds exceeding 85 mm/s, the pump-valve system suffered from instantaneous airflow insufficiency, consequently causing motion trajectory deviations. Although sufficient to validate the control, the tail actuation and control systems are not fully integrated with the quadruped platform, which introduces excessive volume. Furthermore, the bucket-picking demonstration is affected by the system integration and the aging quadruped platform, which added difficulties in the investigation of the soft tail quadruped in performing real world tasks.

To overcome these limitations, it is essential to upgrade the hardware, improve the system integration, and consider inertial forces and load conditions in the PPC under the extended range of motion speeds. We will further investigate the capability of the coordinated movements for real life tasks.

## VII. Conclusion

To realize the non-static and quasi-static motions control of the BVSR tail aiming for implementation on the mobile robotic platform, a pressure predictive control (PPC) based on LSTM network is proposed, which includes IK network, FK network and P-comp network. Based on the same LSTM architecture, the IK, FK and P-comp models are simply obtained by training after one data-acquisition process. To minimize the error between the trajectory predicted by the FK model and the target trajectory, the P-comp model is used to compensate the pressure discrepancy.

For the non-static movements of the BVSR tail, the accuracy of the FK model was firstly verified, which was used as a simulator to perform trajectory calculations. To verify the advantage of the PPC over the IK-only, three typical target trajectories (circle, hexagon and square) were input to PPC and IK-only to calculate the generated pressure respectively. The simulated trajectories of the two methods were compared through FK model (simulator) calculation. In simulation, the average RMSE of the resultant trajectories by PPC are lower than that of IK-only by 69.6% (69.4% in Y-axis, 69.8% in Z-axis coordinates). In experiments, PPC achieves 18% (12% in Y, 24% in Z) reduction in respective RMSE compared to IK-only. The reductions of RMSE proves PPC significantly improves the accuracy in controlling the non-static motions. In following the hexagon and square trajectories with quasi-static motions, the low average RMSE of the hexagon (12.65 mm) and square (12 mm) verify that the PPC can also generate pressures for the quasi-static motion control while trained using non-static data, validating the adaptivity. To evaluate the performance differences between neural network architectures, this study compared the control accuracy of the LSTM-based and GRU-based IK-only controllers on hexagonal and square trajectories. Experimental results indicate that the LSTM structure achieves higher trajectory tracking accuracy, which provides a basis for selecting LSTM as the network foundation in this research. Furthermore, by comparing the hysteresis characteristics of the IK-only and PPC controllers along both the horizontal and vertical directions, it was found that the PPC framework incorporating the P-comp module significantly reduces the hysteresis rate in both motion directions. The study also compared PPC with other control algorithms, and the results demonstrate its clear advantages in both tracking accuracy and the range of motion velocities over which stable accuracy is maintained.

For realizing coordinated motion of the tail in response to the quadruped robot, the real-time performance of the PPC is enhanced by replacing the IK model with IKP, which predicts the next state of the BVSR tail output. The real-time performance is enhanced with the delay time reduced by 60.9%. The coordinated motion performance is investigated by the end-point control experiments, which validates the effectiveness of PPC. The bucket-picking demonstration utilizes the coordinated motion capability and validates the functionality of the soft tail quadruped.